\documentclass[letterpaper, 10 pt, conference]{ieeeconf}

\usepackage{booktabs}
\usepackage{colortbl}
\usepackage{xcolor}
\usepackage{multirow}
\usepackage{array}

\usepackage{amsmath,amsfonts}
\usepackage{array}
\usepackage{enumerate}
\usepackage[utf8]{inputenc}
\usepackage[T1]{fontenc}
\usepackage{amsfonts}
\usepackage{amssymb}
\usepackage{tabularx}
\makeatletter
\let\NAT@parse\undefined
\makeatother
\usepackage{hyperref}
\hypersetup{
    colorlinks=true,
    linkcolor=blue,
    filecolor=magenta,      
    urlcolor=cyan,
    }
\IEEEoverridecommandlockouts 
\usepackage{algorithm}  
\usepackage{algpseudocode}  
\usepackage{breqn}
\usepackage{caption}
\usepackage{subcaption}
\usepackage[dvipsnames]{xcolor}

\usepackage{textcomp}
\usepackage{stfloats}
\usepackage{url}
\usepackage{verbatim}
\usepackage{graphicx}
\usepackage{cite}
\usepackage{color}
\usepackage{colortbl}
\let\labelindent\relax
\usepackage{enumitem}
\usepackage{tabularray}
\usepackage{multirow}
\usepackage{diagbox}
\usepackage{booktabs}
\usepackage{hhline}

\usepackage{framed}
\usepackage[most]{tcolorbox}
\usepackage{xcolor}
\colorlet{shadecolor}{orange!15}
\usepackage[font=footnotesize]{caption}

\definecolor{graytext}{rgb}{0.483, 0.483, 0.4585}       
\definecolor{purpletext}{rgb}{0.626, 0.439, 0.678}      
\definecolor{greentext}{rgb}{0.417, 0.650, 0.294}        
\definecolor{orangetext}{rgb}{0.673, 0.508, 0.228}       
\definecolor{bluetext}{rgb}{0.327, 0.549, 0.7}           
\definecolor{bestcol}{RGB}{219,234,254}   
\definecolor{avggray}{RGB}{245,245,245}  
\definecolor{blacktext}{rgb}{0, 0, 0}
\definecolor{ourblue}{RGB}{219,234,254}    
\definecolor{sectionbg}{RGB}{30,58,138}    
\definecolor{sectionfg}{RGB}{255,255,255}  
\definecolor{avgbg}{RGB}{241,245,249}      
\definecolor{loco}{RGB}{240,249,255}       
\definecolor{ant}{RGB}{240,253,244}        
\definecolor{meta}{RGB}{254,252,232}       
\definecolor{rulecolor}{RGB}{148,163,184}  
\title{\LARGE \bf PrefMoE: Robust Preference Modeling with \\ Mixture-of-Experts Reward Learning}

\author{Ziqin Yuan$^{1}\dag$, Ruiqi Wang$^{1}\dag$, Dezhong Zhao$^{1,2}$, Baijian Yang$^{1}$, and Byung-Cheol Min$^{3}$
\thanks{$\dag$ Equal contribution}
\thanks{$^{1}$Purdue University, West Lafayette, IN, USA.}
\thanks{$^{2}$Beijing University of Chemical Technology, Beijing, China.}
\thanks{$^{3}$Indiana University Bloomington, Bloomington, IN, USA.}
\thanks{Correspondence: \texttt{wang5357@purdue.edu}; \texttt{minb@iu.edu}.}
}

\begin{document}
\setlength{\abovedisplayskip}{1pt} 
\setlength{\belowdisplayskip}{1pt} 

\maketitle

\begin{abstract}
Preference-based reinforcement learning offers a scalable alternative to manual reward engineering by learning reward structures from comparative feedback. However, large-scale preference datasets, whether collected from crowdsourced annotators or generated by synthetic teachers, often contain heterogeneous and partially conflicting supervision, including disagreement across annotators and inconsistency within annotators. Existing reward learning methods typically fit a single reward model to such data, forcing it to average incompatible signals and thereby limiting robustness. To solve this, we propose PrefMoE, a mixture-of-experts reward learning framework for robust preference modeling. PrefMoE learns multiple specialized reward experts and uses trajectory-level soft routing to combine them adaptively, enabling the model to capture diverse latent preference patterns under noisy and heterogeneous preference supervision. A load-balancing regularizer further stabilizes training by preventing expert collapse. Across locomotion benchmarks from D4RL and manipulation tasks from MetaWorld, PrefMoE improves preference prediction robustness and leads to more reliable downstream policy learning than strong single-model baselines. Project page: \texttt{\url{https://sites.google.com/view/moepref}}.
\end{abstract}

\section{Introduction}
Preference-based reinforcement learning (PbRL) has emerged as a promising paradigm for alleviating the challenges of reward engineering in robotics~\cite{christiano2017deep}. Rather than relying on manually specified reward functions, PbRL learns reward models directly from human judgments~\cite{wang2022feedback,ibarz2018reward,wang2025personalization}, typically through pairwise comparisons of trajectory segments that define a principled learning objective~\cite{wirth2017survey}. Despite its effectiveness, a key bottleneck is that collecting sufficient preference labels from humans is always costly.
\begin{figure}[t]
    \centering
    \includegraphics[width=0.70\columnwidth]{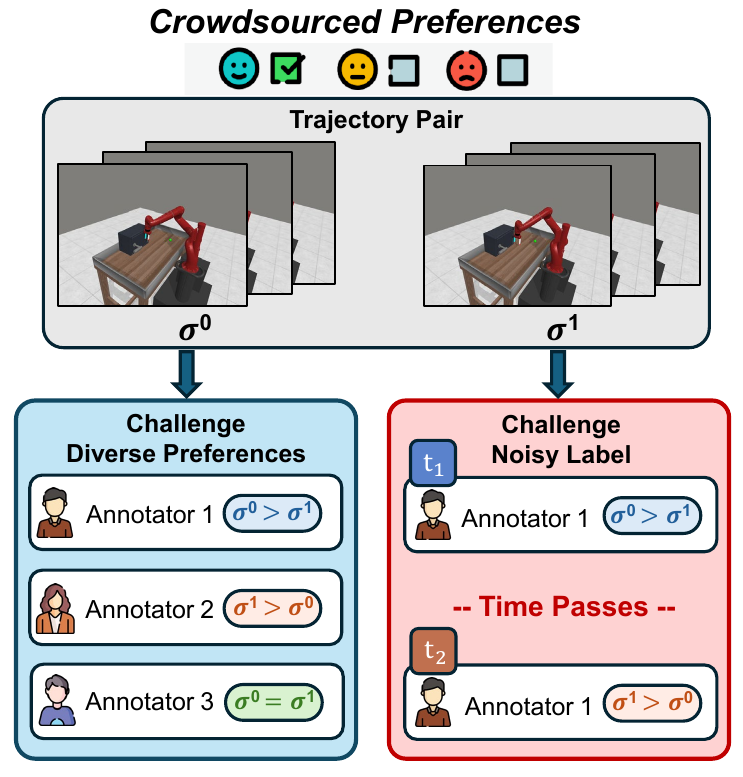}
    \vspace{-5pt}
    \caption{\textbf{Two key challenges in preference-based reward learning.} Given the same robot trajectory pair $(\sigma^A, \sigma^B)$ (\textbf{left}), crowdsourced annotators may produce conflicting labels, e.g., one preferring $\sigma^A \succ \sigma^B$, another $\sigma^B \succ \sigma^A$, another indicating equal preference, and another providing no label (\textbf{left}, \textbf{\textit{inter-annotator disagreement}}). Moreover, the same annotator may label the identical pair differently at time $t_1$ versus $t_2$ (\textbf{right}, \textbf{\textit{intra-annotator inconsistency}}). Both effects introduce noise into the preference pool, motivating the need for a robust reward model that can handle diverse and inconsistent human feedback.}
    \label{fig:challenges}
    \vspace{-15pt}
\end{figure}

To mitigate this limitation, recent work has explored crowdsourcing-based pipelines and infrastructure for large-scale preference collection~\cite{ouyang2022traininglanguagemodelsfollow}. For example, UniRLHF~\cite{yuan2024unirlhf} introduces a universal annotation platform and large-scale datasets spanning diverse environments and user populations, significantly scaling preference data collection. However, such large-scale annotation also introduces substantial label variability. In practice, annotators often rely on different latent criteria, and even the same annotator may produce inconsistent judgments over time. As illustrated in Fig.~\ref{fig:challenges}, this results in both inter-annotator conflicts and intra-annotator inconsistency.

More recent work has also explored synthetic preference supervision. For example, foundation models such as large language models (LLMs) have been used as synthetic teachers to generate preference labels or critiques for reinforcement learning~\cite{wang2025prefclm,wang2024rl,wang2025primt}. While synthetic feedback provides a scalable alternative to human annotation, it similarly introduces heterogeneity and uncertainty, as model-generated judgments may follow implicit criteria that differ from human intent or vary across prompts and contexts.

Consequently, large-scale preference datasets, whether collected from crowdsourced human annotators or generated by LLMs, often contain heterogeneous and partially conflicting supervision signals. However, existing reward learning methods in PbRL~\cite{zhao2025prefmmt,lee2021pebble,kim2023preference_transformer} typically fit a single reward model to such data, implicitly assuming a unified preference structure. As a result, the learned reward may average incompatible criteria and fail to generalize across diverse scenarios.

To address this limitation, we propose PrefMoE, a mixture-of-experts (MoE) framework for reward learning under heterogeneous preference supervision. Instead of fitting a single reward model, PrefMoE learns multiple specialized reward experts implemented as reward heads on top of a multimodal transformer structure, and uses a gating mechanism to dynamically combine them for each trajectory (Fig.~\ref{fig:framework}). By modeling diverse latent preference patterns explicitly, PrefMoE improves robustness to noisy and conflicting annotations and leads to more reliable reward estimation.

Our key contributions can be summarized as follows:
\begin{itemize}[leftmargin=*]
    \item We study preference reward learning under realistic large-scale settings where the available preference pool is heterogeneous and noisy, arising from crowdsourced human feedback and synthetic supervision.

    \item We propose PrefMoE, a MoE multimodal transformer reward model that explicitly captures heterogeneous preference patterns via trajectory-level expert routing.

    \item Through extensive experiments on locomotion benchmarks from D4RL~\cite{fu2020d4rl} and manipulation tasks from MetaWorld~\cite{yu2020metaworld}, we show that PrefMoE significantly improves preference reward robustness and yields more reliable downstream policy learning than state-of-the-art reward modeling baselines.
\end{itemize}

\section{Background and Preliminary}

\subsection{Preference-based Reinforcement Learning}

In PbRL, a reward model $r_\psi$ is learned from comparative feedback over trajectory segments. Given two segments, $\sigma^0$ and $\sigma^1$, where each segment is $\sigma=\{(s_t,a_t)\}_{t=1}^T$, an annotator provides a preference label $y \in \{0,1,0.5\}$ indicating whether $\sigma^0$ or $\sigma^1$ is preferred, or whether the two are tied.

Following prior work, preferences are modeled with the Bradley--Terry (BT) model~\cite{bradley1952rank}, where the probability of preferring $\sigma^1$ over $\sigma^0$ is
\begin{equation}
P[\sigma^1 \succ \sigma^0; \psi] =
\frac{\exp\left(\sum_{t=1}^T r_\psi(s_t^1,a_t^1)\right)}
{\sum_{j \in \{0,1\}} \exp\left(\sum_{t=1}^T r_\psi(s_t^j,a_t^j)\right)}.
\end{equation}
The reward model is trained by minimizing the cross-entropy loss over a preference dataset $\mathcal{D}$:
\begin{equation}
\begin{split}
\mathcal{L}_{\text{BT}}(\psi)
= - \mathbb{E}_{(\sigma^0,\sigma^1,y)\sim\mathcal{D}} \Big[
& y \log P[\sigma^1 \succ \sigma^0] \\
& + (1-y)\log P[\sigma^0 \succ \sigma^1]
\Big].
\end{split}
\label{bt}
\end{equation}

\subsection{Preference Modeling in PbRL}
Early PbRL methods typically adopt a Markovian reward formulation, predicting rewards from the current state-action pair only. However, human preferences are often non-Markovian, depending on temporal context and trajectory-level consistency rather than isolated transitions~\cite{casper2023open,kim2023preference_transformer}. Motivated by this, recent work has shifted toward sequence-based reward modeling, where transformers are used to capture long-range dependencies in preference judgments~\cite{kim2023preference_transformer}.

Building on sequence modeling, more recent work further shows that robot preference prediction is inherently multimodal. In particular, PrefMMT~\cite{zhao2025prefmmt} improves reward learning by explicitly modeling both intra-modal temporal dynamics and inter-modal state-action interplays. However, despite these architectural advances, existing approaches still typically fit a single reward model, implicitly assuming that preference labels come from a coherent and consistent source. This assumption is often violated in realistic large-scale preference pools collected from crowdsourcing or synthetic supervision~\cite{yuan2024unirlhf,wang2025prefclm}.

Recent work has also explored robustness to corrupted preference labels. For example, RIME~\cite{cheng2024rime} introduces a denoising discriminator to filter noisy labels during training. While effective for handling instance-level corruption, such methods primarily treat preference unreliability as label noise and do not address the more structured heterogeneity that arises when different annotators apply different latent criteria. Moreover, they rely on active environment interaction and are not directly applicable to the offline PbRL setting, where the preference pool is fixed in advance. 

In contrast, we explicitly model heterogeneous preference supervision at the reward-learning level. We introduce a mixture-of-experts routing mechanism. While we implement this atop a multimodal transformer backbone, the routing specifically occurs at the inter-modal fusion stage to capture diverse human evaluation criteria.

\subsection{Mixture-of-Experts for Preference Modeling}

MoE is a natural fit for preference learning under heterogeneous supervision, where different annotators or synthetic teachers may implicitly apply different evaluation criteria. Rather than forcing a single reward model to average these partially conflicting signals, PrefMoE uses MoE to enable conditional specialization across latent preference patterns.

Our use of MoE is distinct from standard applications that place experts in generic feed-forward layers~\cite{puigcerver2024from}. Instead, PrefMoE introduces experts as attentive heads at the inter-modal fusion stage of a multimodal transformer reward model. Each expert is an inter-modal encoder with its own cross-attention parameters, allowing different experts to specialize in different state-action interaction patterns when evaluating trajectory quality. A trajectory-level gating network then softly combines expert outputs into the final reward estimate based on the context of trajectories.

\begin{figure*}[t]
    \centering
    \includegraphics[width=\textwidth]{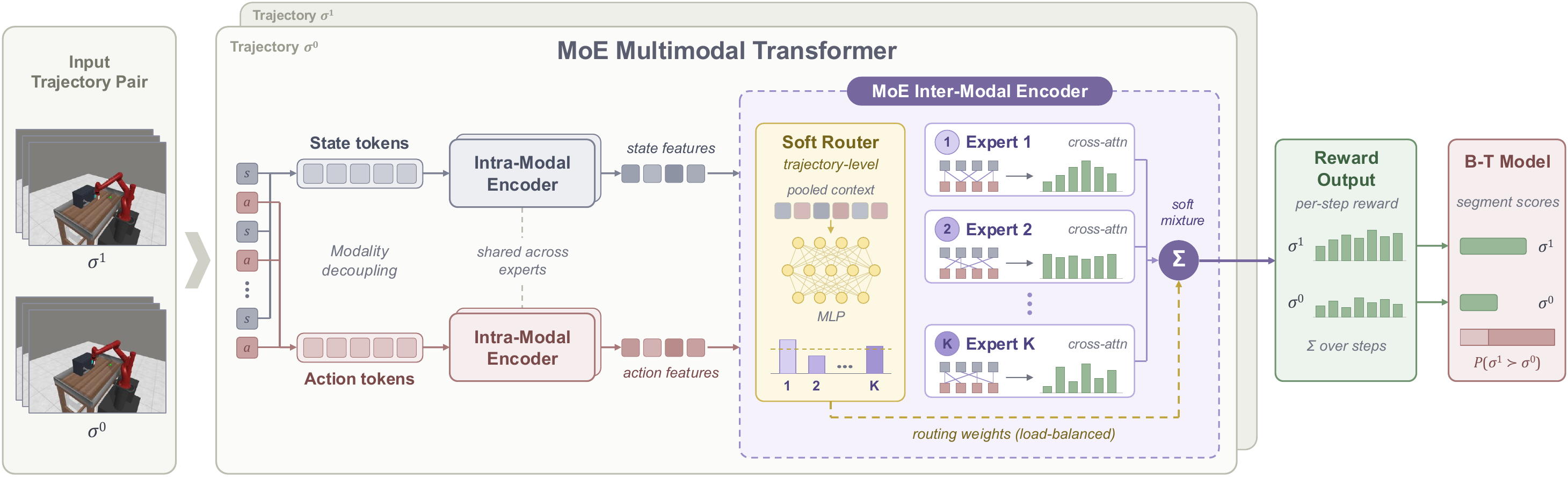}
    \vspace{-15pt}
    \caption{\textbf{Overview of PrefMoE.} A trajectory $\sigma$ is first decoupled into state and action streams, which are independently processed by shared intra-modal encoders. The resulting representations are pooled into a context vector, from which a two-layer MLP soft router produces $K$ routing weights. Each of the $K$ expert inter-modal encoders computes a state--action cross-attention reward sequence; the final per-step reward is a convex mixture; these are aggregated into segment score $\rho^{\text{MoE}}(\sigma)$, which feeds the Bradley--Terry preference predictor.}
    \vspace{-5pt}
    \label{fig:framework}
\end{figure*}

\section{Methodology}

\subsection{Problem Formulation}
We assume access to a preference dataset $\mathcal{D}=\{(\sigma_i^0,\sigma_i^1,y_i)\}_{i=1}^{|\mathcal{D}|}$ collected from a diverse pool of annotators or synthetic teachers. Our goal is to learn a reward model $\hat{r}_\psi$ that remains reliable under preference-pool diversity and label noise.

\subsection{Overview of PrefMoE}
To address heterogeneous preference supervision, we propose PrefMoE, a MoE multimodal transformer reward model as shown in Fig.~\ref{fig:framework}. Following~\cite{zhao2025prefmmt}, each trajectory is processed in two stages. First, shared intra-modal encoders model temporal dynamics within the state and action streams. Second, inter-modal encoders integrate the two streams to evaluate trajectory quality.

We introduce MoE only at the inter-modal stage. The reason is that intra-modal encoders mainly capture temporal structures influencing human judgments, including salient progress cues such as whether key events or task completion are achieved, which are largely governed by environment dynamics and agent behavior and thus remain relatively consistent across annotators~\cite{zhao2025prefmmt}. By contrast, preference heterogeneity primarily arises in how state and action information are jointly interpreted to assess trajectory quality. Inter-modal interactions reflect more subtle and subjective criteria, such as smoothness, efficiency, and other nuanced aspects of execution, which are more likely to vary across annotators and to be poorly represented by a single shared model. Placing MoE at this stage therefore allows PrefMoE to specialize different experts for different latent evaluation criteria while maintaining shared temporal representations for each modality.

\subsubsection{Shared Backbone}
A trajectory $\sigma$ is represented as aligned state and action streams, written as $s_{1:T}$ and $a_{1:T}$. Each stream is embedded and augmented with shared temporal positional encodings:
\begin{align}
\mathbf{x}^S &= f^S_e(s_{1:T}) + \mathbf{E}_{1:T} \in \mathbb{R}^{T \times d}, \\
\mathbf{x}^A &= f^A_e(a_{1:T}) + \mathbf{E}_{1:T} \in \mathbb{R}^{T \times d},
\end{align}
where $f^S_e$ and $f^A_e$ are modality-specific embedding layers and $\mathbf{E}_{1:T}$ denotes shared positional encodings. Each stream is then processed by a shared intra-modal encoder to produce temporally enriched representations $\tilde{\mathbf{x}}^S, \tilde{\mathbf{x}}^A \in \mathbb{R}^{T \times d}$. These shared representations provide a common temporal backbone across all experts and improve sample efficiency by avoiding redundant modeling of modality-specific dynamics.

\subsubsection{Routing}
PrefMoE performs routing at the \emph{trajectory level} to determine how inter-modal interactions should be evaluated. This choice is consistent with the supervision granularity in PbRL, since preference labels are provided for entire trajectory segments rather than individual transitions. Moreover, trajectory-level routing allows expert selection to depend on full temporal context rather than local observations.

To obtain a routing signal, we compute a global context vector $\mathbf{c}(\sigma)\in\mathbb{R}^{d_r}$ by pooling the intra-modal representations:
\begin{equation}
\mathbf{c}(\sigma) = \text{LayerNorm}\left(W_c \cdot \frac{1}{T}\sum_{t=1}^T (\tilde{\mathbf{x}}^S_t + \tilde{\mathbf{x}}^A_t)\right).
\end{equation}
A two-layer MLP then produces routing weights over $K$ experts:
\begin{equation}
\mathbf{g}(\sigma) = \text{softmax}\!\left(W_{g_2}\,\text{ReLU}(W_{g_1}\mathbf{c}(\sigma)+\mathbf{b}_{g_1})+\mathbf{b}_{g_2}\right)\in\mathbb{R}^K.
\end{equation}
We use \emph{soft routing}, so that all experts contribute with continuous weights. This preserves end-to-end differentiability and allows PrefMoE to represent mixed or ambiguous latent preference criteria within a single trajectory.

\subsubsection{Expert Encoders}
Each expert $\mathcal{E}_k$ is an inter-modal encoder with its own cross-attention parameters, enabling expert-specific modeling of state--action interaction patterns. Concretely, expert $k$ applies two causal cross-attention pathways, from state to action and from action to state:
\begin{align}
\mathbf{z}^{S,k} &= \text{softmax}\!\left(\frac{\mathbf{Q}^{S,k}(\mathbf{K}^{A,k})^\top}{\sqrt{d^A_k}} + \mathbf{M}\right)\mathbf{V}^{A,k},\\
\mathbf{z}^{A,k} &= \text{softmax}\!\left(\frac{\mathbf{Q}^{A,k}(\mathbf{K}^{S,k})^\top}{\sqrt{d^S_k}} + \mathbf{M}\right)\mathbf{V}^{S,k},
\end{align}
where $\mathbf{M}\in\mathbb{R}^{T\times T}$ is the causal mask. Intuitively, these expert-specific inter-modal encoders allow different experts to specialize in different latent evaluation strategies when judging trajectory quality. Each expert outputs a per-step reward sequence $\hat{\mathbf{r}}^k_{1:T}\in\mathbb{R}^{T}$ and the corresponding segment score
\begin{equation}
\rho^k(\sigma)=\sum_{t=1}^T \hat{r}^k_t(\sigma).
\end{equation}

\subsubsection{Mixture Reward}
The final per-step reward is obtained as a convex mixture of expert rewards:
\begin{equation}
\hat{r}_t(\sigma) = \sum_{k=1}^K g_k(\sigma)\,\hat{r}^k_t(\sigma),
\end{equation}
and the resulting segment score is
\begin{equation}
\rho^{\text{MoE}}(\sigma)=\sum_{k=1}^K g_k(\sigma)\,\rho^k(\sigma).
\end{equation}
This mixture formulation enables PrefMoE to interpolate among latent preference criteria while remaining fully compatible with the BT preference modeling framework in PbRL.

\subsection{Training Objective}
PrefMoE is trained end-to-end with the trajectory-level BT loss as defined in Eq.~\ref{bt}. To prevent expert collapse and encourage utilization of diverse experts, we add a load-balancing regularizer over a mini-batch $\mathcal{B}$:
\begin{equation}
\mathcal{L}_{\text{bal}}(\psi;\mathcal{B}) = \sum_{k=1}^K \left(\frac{1}{|\mathcal{B}|}\sum_{\sigma\in\mathcal{B}} g_k(\sigma) - \frac{1}{K}\right)^2.
\end{equation}
The final training objective is
\begin{equation}
\mathcal{L}(\psi)=\mathcal{L}_{\text{BT}}(\psi)+\lambda\,\mathcal{L}_{\text{bal}}(\psi),
\end{equation}
where $\lambda$ controls the strength of expert balancing.

\subsection{Inference for RL Training}
During downstream policy optimization, PrefMoE is applied to a sliding window of the most recent $H$ transitions to produce a non-Markovian reward estimate consistent with the segment-level preference formulation used during training. The routing network is evaluated once per window to obtain $\mathbf{g}(\cdot)$, and the final reward is computed as the soft mixture of expert outputs. This allows the learned preference model to provide context-dependent reward signals for the downstream RL algorithm while preserving the trajectory-level specialization learned from heterogeneous preference supervision.
\section{Experimental Setups}

\subsection{Environments and Preference Labels}
We evaluate on three benchmark domains: i) six D4RL locomotion tasks: halfcheetah, walker2d, and hopper in medium-expert and medium-replay variants; ii) two AntMaze navigation tasks with sparse binary rewards; and iii) five MetaWorld manipulation tasks. Locomotion and AntMaze performance is reported as D4RL normalized score~\cite{fu2020d4rl}; MetaWorld performance is reported as task success rate.

Each preference query presents a pair of trajectory segments $(\sigma^0, \sigma^1)$ and an annotator label $y \in \{0, 1, 0.5\}$ indicating preference. We use crowdsourced labels from~\cite{yuan2024unirlhf} for locomotion and AntMaze, and labels from~\cite{zhao2025prefmmt} for MetaWorld.

\subsection{Baselines}

\noindent \textbf{MR~\cite{lee2021pebble}.}
A traditional Markovian reward model that scores each state--action pair independently using an MLP. This represents the most common reward modeling approach in PbRL.

\noindent \textbf{PrefMMT~\cite{zhao2025prefmmt}.}
A multimodal transformer-based reward model that processes trajectories through intra-modal encoders and an inter-modal cross-attention encoder. PrefMMT represents the current state-of-the-art sequence-based preference modeling architecture.

\noindent \textbf{RIME-offline~\cite{cheng2024rime}.}
An adaptation of RIME to the offline setting, where preference labels are fixed rather than collected online. RIME employs an ensemble-based trust filtering mechanism to mitigate noisy annotations, representing approaches that address preference noise through label filtering.

\subsection{Training Settings}

Following~\cite{zhao2025prefmmt}, we adopt the offline PbRL setting so that all methods are trained and evaluated on the same fixed human preference dataset, ensuring fair comparison without confounding differences in data collection. The training involves several steps:

\begin{enumerate}[leftmargin=*]
  \item \textbf{Reward model training.}
        Train each reward model on the human preference dataset to obtain $\hat{r}_\psi$ with different reward learing methods.

  \item \textbf{Reward relabeling.}
        Apply $\hat{r}_\psi$ to all transitions in the offline RL buffers to assign rewards.

  \item \textbf{Policy training.}
        Train a policy on the relabeled buffers using Implicit Q-Learning (IQL)~\cite{kostrikov2021iql}. All hyperparameters follow the original benchmark configurations.
\end{enumerate}

\subsection{Evaluation Protocol}

For D4RL locomotion and AntMaze tasks, we report the normalized score as defined in the D4RL benchmark~\cite{fu2020d4rl}:

$\text{normalized score} = 100 \times
\frac{\text{score}-\text{random score}}
     {\text{expert score}-\text{random score}}$

For locomotion tasks, we average over 10 evaluation episodes every 5{,}000 policy gradient steps. For AntMaze tasks, we average over 100 evaluation episodes every 100{,}000 gradient steps to account for the longer episode horizon
and sparser reward signal. For MetaWorld tasks, we report the binary task success rate provided natively by the environment, averaged over 10 evaluation episodes every 5{,}000 gradient steps. All results are reported as the mean $\pm$ standard deviation over five independent random seeds.

\subsection{Implementation Details}

\noindent \textbf{PrefMoE architecture.}
The shared backbone is a GPT-2-style transformer. The MoE component uses $K\!=\!4$ expert inter-modal encoders with independent parameters, a routing dimension $d_r\!=\!128$, and load-balancing coefficient $\lambda\!=\!10^{-2}$.

\noindent \textbf{Reward model training.}
All reward models are trained with the AdamW optimizer at a peak learning rate of $10^{-4}$ with cosine decay scheduling, batch
size 256, for 10{,}000 epochs. The preference dataset is split 90/10 into training and validation sets; the model checkpoint with the lowest validation loss is used for reward relabeling. Training is performed on a single NVIDIA 5090 GPU.

\noindent \textbf{RIME-offline training.}
RIME-offline uses an ensemble of three MLPs, each with three hidden layers of dimension 256 and LeakyReLU activations, and a final Tanh activation on the scalar output. All three members are optimised jointly with Adam at a learning rate of $3\!\times\!10^{-4}$. The learning rate is linearly warmed up from zero to its peak over the first 3{,}000 epochs; the KL trust-filter tightness parameter $\beta$ decays linearly from 3.0 to 1.0 over the same interval. Samples whose ensemble KL divergence exceeds $-\log(0.001)\approx 6.9$ have their labels temporarily flipped for that epoch's gradient updates and then reverted; samples below the adaptive trust threshold are excluded from updates entirely. Training uses 2{,}000 preference pairs of segment length $T\!=\!200$, a mini-batch size of 256, and runs for 10{,}000 epochs. The checkpoint with the lowest held-out cross-entropy loss is retained for reward relabeling, following the same protocol as all other baselines.

\noindent \textbf{$K$ ablation (number of experts).}
To assess the effect of expert count, we vary $K \in \{1, 2, 4\}$ while keeping all other hyperparameters fixed. Setting $K{=}1$ reduces PrefMoE exactly to PrefMMT, providing a direct controlled comparison without any code changes.

\noindent \textbf{Noise robustness ablation.}
To evaluate model robustness to annotation noise, we apply additional synthetic label flips at rates $\Delta p \in \{+0.1, +0.2, +0.3\}$ on top of the natural 100-annotator label noise, simulating increasingly degraded annotation quality across D4RL locomotion tasks. For the annotator-count ablation, we subsample $N_{\text{ann}} \in \{10, 25, 50, 100\}$ annotators from the full human label pool, keeping all other settings fixed.

\noindent \textbf{Annotator-pool-size ablation.}
To evaluate how each method responds to growing annotation diversity, we subsample $N_{\text{ann}} \in \{10, 25, 50, 100\}$ annotators from the full crowdsourced label pool and retrain all reward models under each setting. A smaller pool produces more homogeneous labels with lower inter-annotator disagreement; a larger pool introduces greater preference diversity and natural label noise. No additional synthetic noise is applied ($\Delta p\!=\!+0$), and all other hyperparameters are kept fixed, isolating annotator-pool diversity as the sole variable.

\section{Results and Analysis}

\subsection{Comparison with Baselines}

Table~\ref{tab:table1_prefmoe} presents downstream policy performance across three benchmark domains. Overall, PrefMoE achieves the highest average score in every domain. However, the task-level picture is more nuanced: no single method dominates uniformly, and the relative ordering shows substantial variance depending on the interplay between dataset quality, trajectory length, and annotation noise.

\begin{table*}[!t]
\centering
\caption{\textbf{Comparison of PrefMoE with baselines on D4RL and MetaWorld.}
Locomotion and AntMaze labels are crowdsourced from \textbf{100 annotators};
MetaWorld labels use \textbf{10 annotators}.
Locomotion and AntMaze report D4RL normalized score;
MetaWorld reports task success rate~(\%).
$K\!=\!4$ for all PrefMoE results.
Values are mean $\pm$ std over five seeds.
\textbf{Bold} + \colorbox{bestcol}{\strut shading}: best per task.}
\label{tab:table1_prefmoe}
\vspace{-2mm}
\fontsize{8.5}{11}\selectfont
\renewcommand{\arraystretch}{1.1}
\begin{tabular*}{\textwidth}{@{\extracolsep{\fill}}
  >{\raggedright\arraybackslash}p{4.4cm}
  cccc @{}}
\toprule
\textbf{Task}
  & \textbf{MR}
  & \textbf{RIME-offline}
  & \textbf{PrefMMT}
  & \textbf{PrefMoE (Ours)} \\
\midrule

\multicolumn{5}{@{}l}{\textit{D4RL Locomotion (Normalized Score)}~$\uparrow$} \\[1pt]
halfcheetah-medium-expert-v2 & $69.9{\pm}7.4$ & $70.1{\pm}5.9$ & $79.8{\pm}5.0$ & \cellcolor{bestcol}$\mathbf{83.7{\pm}1.5}$ \\
halfcheetah-medium-replay-v2 & $35.0{\pm}7.9$ & $42.4{\pm}6.1$ & $47.0{\pm}5.3$ & \cellcolor{bestcol}$\mathbf{56.3{\pm}4.1}$ \\
walker2d-medium-expert-v2    & $96.4{\pm}2.9$ & $96.5{\pm}3.4$ & $107.2{\pm}2.4$ & \cellcolor{bestcol}$\mathbf{108.4{\pm}0.7}$ \\
walker2d-medium-replay-v2    & $68.3{\pm}4.6$ & $72.8{\pm}13.7$ & $70.4{\pm}2.1$ & \cellcolor{bestcol}$\mathbf{79.1{\pm}5.4}$ \\
hopper-medium-expert-v2      & $67.0{\pm}5.7$ & $77.3{\pm}4.8$ & $81.4{\pm}6.5$ & \cellcolor{bestcol}$\mathbf{87.6{\pm}4.3}$ \\
hopper-medium-replay-v2      & $27.5{\pm}4.5$ & $44.8{\pm}9.3$ & $78.6{\pm}3.2$ & \cellcolor{bestcol}$\mathbf{85.8{\pm}2.1}$ \\
\rowcolor{avggray}
\textit{Gym-Average}         & \textit{$60.7{\pm}5.2$} & \textit{$67.3{\pm}7.2$} & \textit{$77.4{\pm}3.8$} & \cellcolor{bestcol}\textit{$\mathbf{84.3{\pm}3.0}$} \\

\midrule
\multicolumn{5}{@{}l}{\textit{AntMaze (Normalized Score)}~$\uparrow$} \\[1pt]
antmaze-large-play-v2        & $6.4{\pm}1.5$  & $10.4{\pm}2.3$ & $36.8{\pm}2.4$ & \cellcolor{bestcol}$\mathbf{44.8{\pm}2.2}$ \\
antmaze-medium-play-v2       & $44.8{\pm}4.8$ & $51.2{\pm}4.5$ & $60.4{\pm}3.4$ & \cellcolor{bestcol}$\mathbf{65.1{\pm}3.1}$ \\
\rowcolor{avggray}
\textit{AntMaze-Average}     & \textit{$25.6{\pm}3.2$} & \textit{$30.8{\pm}3.4$} & \textit{$48.6{\pm}3.0$} & \cellcolor{bestcol}\textit{$\mathbf{55.0{\pm}2.7}$} \\

\midrule
\multicolumn{5}{@{}l}{\textit{MetaWorld (Success Rate, \%)}~$\uparrow$} \\[1pt]
button-press-v2              & $66.4{\pm}0.9$ & $63.8{\pm}4.2$ & $78.6{\pm}1.5$ & \cellcolor{bestcol}$\mathbf{81.4{\pm}2.3}$ \\
door-open-v2                 & $64.7{\pm}4.8$ & $68.6{\pm}5.3$ & \cellcolor{bestcol}$\mathbf{80.8{\pm}4.1}$ & $74.7{\pm}4.4$ \\
drawer-close-v2              & $60.3{\pm}4.2$ & $63.1{\pm}3.8$ & $74.8{\pm}3.7$ & \cellcolor{bestcol}$\mathbf{78.5{\pm}2.9}$ \\
window-close-v2              & \cellcolor{bestcol}$\mathbf{75.4{\pm}1.2}$ & $72.8{\pm}2.6$ & $74.4{\pm}2.1$ & $73.8{\pm}2.4$ \\
sweep-into-v2                & $46.7{\pm}1.2$ & $44.2{\pm}5.8$ & $58.2{\pm}1.9$ & \cellcolor{bestcol}$\mathbf{64.5{\pm}3.1}$ \\
\rowcolor{avggray}
\textit{MetaWorld-Average}   & \textit{$62.7{\pm}2.5$} & \textit{$62.5{\pm}4.3$} & \textit{$73.4{\pm}2.7$} & \cellcolor{bestcol}\textit{$\mathbf{74.6{\pm}3.0}$} \\

\bottomrule
\end{tabular*}
\vspace{-4mm}
\end{table*}

\noindent \textbf{D4RL Locomotion.}
The locomotion results organize naturally around two dataset regimes with opposite implications for preference model design.

On replay-buffer tasks, the offline dataset spans a continuous quality gradient from near-random exploration to near-expert behavior, and the preference signal between two segments is often carried by trajectory-level patterns rather than by any individual step.
A burst of high joint velocity may appear locally rewarding yet precede a fall, and whether a full segment sustains controlled locomotion is invisible to a per-step reward model.
Both MR and RIME-offline are constrained to this Markovian view and plateau well below PrefMMT and PrefMoE on all replay tasks.
RIME-offline's denoising raises performance over MR, indicating that crowdsourced labels on mixed-quality data are noisy, but the Markovian constraint limits how much denoising alone can recover.

The expert tasks present a more homogeneous dataset, yet temporal models continue to lead across all three expert splits.
When all demonstrations originate from a near-expert policy, adjacent segments differ primarily in fine-grained sequential patterns such as joint coordination, velocity consistency, and sustained balance across the full trajectory.
These cues span multiple timesteps and are invisible to any per-step reward model, explaining why PrefMoE and PrefMMT outperform Markovian baselines even here.
The advantage of PrefMoE over PrefMMT is modest on expert tasks relative to replay splits, reflecting that expert preference pools have lower inter-annotator diversity: with most trajectories near-optimal, annotators converge on similar evaluation criteria and the router tends toward near-uniform weighting across experts.
RIME-offline's trust filter, designed to discard corrupted labels, has the opposite effect on clean expert data: it prunes informative near-boundary samples rather than corrupted ones, reducing the effective training set and lowering performance on most expert tasks.
Notably, on walker2d-medium-expert-v2, both PrefMMT and PrefMoE surpass the oracle IQL score obtained with ground-truth task reward, suggesting that preference-based reward models can encode behavioral understanding that complements or exceeds hand-designed task rewards in sufficiently structured environments.

\noindent \textbf{AntMaze.}
AntMaze exposes the structural limitation of Markovian reward modeling more sharply than any locomotion task.
In these sparse-reward environments, the only signal available at preference query time is whether a trajectory reaches the goal; whether a given state-action pair is preferable depends entirely on the long-range context of the agent's path, including the direction it is heading and how efficiently it traverses the maze structure.
No amount of label denoising can recover temporal context that was never observed, so MR and RIME-offline converge to near-zero performance on antmaze-large-play-v2 regardless of label quality, while PrefMMT and PrefMoE maintain a substantial advantage.

On antmaze-medium-play-v2, PrefMMT narrowly outperforms PrefMoE.
The medium maze's simpler topology reduces the diversity of navigational sub-tasks present in the preference pool: with fewer distinct path-level strategies for annotators to evaluate differently, the soft router tends toward a near-uniform weighting across experts.
In this limit the additional parameterization of MoE provides no benefit over a single well-trained inter-modal encoder, and PrefMMT is sufficient.
This reversal is consistent with the central prediction of the MoE design, that routing earns its advantage only when genuine latent diversity exists in the annotator preference pool.

\noindent \textbf{MetaWorld.}
The manipulation results suggest that the benefit of MoE routing scales with the structural complexity of the preference signal rather than with task difficulty in isolation. PrefMoE leads on button-press-v2, drawer-close-v2, and sweep-into-v2. Each task requires a multi-phase behavioral sequence spanning approach, contact initiation, and sustained force or object displacement.
Annotators evaluating such sequences naturally weight the phases differently according to their latent criteria, and this inter-annotator diversity is precisely the signal the routing mechanism exploits: different expert inter-modal encoders specialize in different evaluation strategies, and the soft mixture aggregates them coherently.
Sweep-into-v2 shows the highest variance across all MetaWorld tasks, reflecting the stochastic nature of contact-rich relocation and the resulting annotator disagreement on near-threshold episodes; the MoE architecture is particularly suited to such settings because routing assignments adapt to the evaluation style implied by each incoming preference pair.

On door-open-v2, PrefMMT achieves the highest score, with PrefMoE below it.
The temporal context of the full reaching trajectory, including approach angle, gripper speed, and pre-contact positioning, provides richer preference signal than per-step features alone, giving PrefMMT a clear advantage over Markovian baselines.
PrefMoE does not improve further over PrefMMT on this task: the critical preference signal is the contact event itself, a shared criterion across annotators, and with little latent diversity in how evaluators weight the trajectory the soft router converges toward near-uniform expert weighting, providing no representational benefit beyond a single well-trained inter-modal encoder.
window-close-v2 is a more extreme case: a fully stereotyped reaching-grasping-pushing sequence with near-zero inter-annotator disagreement, where MR achieves the highest score across all methods and neither transformer model adds value over the per-step baseline.
These two tasks jointly suggest that MoE routing is most valuable when annotator criteria genuinely diverge across preference pairs, a condition that single-contact and stereotyped tasks do not satisfy.

RIME-offline does not consistently improve over MR in MetaWorld.
On most tasks, RIME-offline scores below MR despite its denoising mechanism.
The trust filter identifies corrupted labels through ensemble KL divergence, but crowdsourced annotations on these structured manipulation tasks tend to be internally consistent rather than randomly corrupted.
Samples flagged as high-KL often reflect genuine preference ambiguity near task-phase boundaries rather than label errors, and pruning them reduces the effective training set without correcting any actual corruption.
In this offline adaptation, the mechanism that is designed to improve robustness inadvertently discards informative signal, leaving a simple per-step baseline as the stronger choice on tasks where the preference label quality is inherently high.

\subsection{Ablation Studies}

\subsubsection{Does Increasing the Number of Experts $K$ Yield Monotonic Gains, and Does Annotation Diversity Drive Them}

\begin{figure}[t]
  \centering
  \includegraphics[width=\columnwidth]{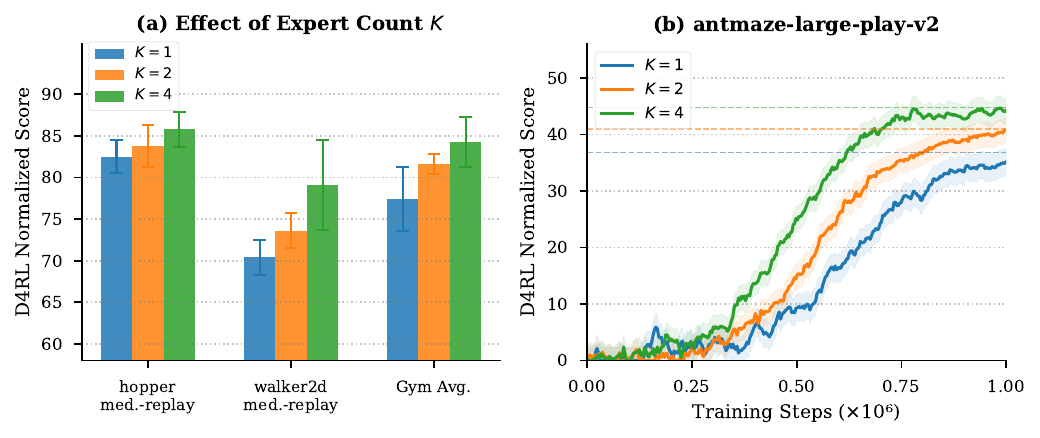}
  \caption{\textbf{Effect of expert count $K$.} \textbf{(a)}~D4RL Gym-average score for $K \in \{1, 2, 4\}$ with $\pm$std error bars over five seeds. \textbf{(b)}~RL policy learning curves on antmaze-large-play-v2, the benchmark with the highest annotator diversity; dashed horizontals mark the table mean for each $K$. Increasing $K$ accelerates goal discovery and raises the final score, indicating that the MoE routing captures distinct navigational evaluation modes that a single encoder must compromise between.}
  \label{fig:k_ablation}
\end{figure}

We vary $K \in \{1, 2, 4\}$ while fixing all other hyperparameters. By construction, $K\!=\!1$ reduces PrefMoE to PrefMMT (the single expert recovers the PrefMMT inter-modal encoder and $\mathcal{L}_{\text{bal}}$ vanishes), providing a direct controlled comparison without any implementation change. All experiments use the 100-annotator crowdsourced label pool and no synthetic noise ($\Delta p\!=\!+0$), identical to the main table setting.
Fig.~\ref{fig:k_ablation} visualises both the aggregate locomotion performance and the downstream policy behaviour on the most challenging AntMaze task.

Under the 100-annotator setting, performance improves monotonically with $K$, with a substantial gain from $K\!=\!1$ to $K\!=\!2$ at $+4.2$ and a further increment from $K\!=\!2$ to $K\!=\!4$ at $+2.7$, for a total improvement of $+6.9$ from PrefMMT to PrefMoE.
These gains are considerably larger than those observed in smaller, more homogeneous annotation pools, indicating that the MoE extension's benefit scales directly with the diversity of the annotator pool. The $K\!=\!1\!\to\!2$ jump indicates that a second expert captures a qualitatively distinct preference mode; for hopper-replay, one expert appears to specialize in gait stability while the other focuses on forward progress, competing preferences that a single encoder must compromise between. Subsequent experts provide finer-grained specialization, and the three-step $K\!=\!1\!\to\!4$ jump captures most of the recoverable diversity within a 2{,}000-pair preference pool. On antmaze-large-play-v2 (Fig.~\ref{fig:k_ablation}b), increasing $K$ accelerates goal discovery and raises the plateau: the K=1 policy converges to $36.8$ while $K\!=\!4$ reaches $44.8$, an $8$ gain on the most diversity-rich navigation task in our benchmark. The load-balancing regularizer $\lambda\mathcal{L}_{\text{bal}}$ is essential: without it, ablations show routing collapse to a single dominant expert within $1{,}000$ training epochs, eliminating any representational benefit.

\subsubsection{Does PrefMoE's Routing Provide Implicit Noise Resilience Without Explicit Denoising}

We inject additional synthetic label flips at rates $\Delta p \in \{+0.1,\,+0.2,\,+0.3\}$ on top of the natural heterogeneous noise already present in the 100-annotator pool. $\Delta p\!=\!+0$ denotes the unperturbed baseline (natural crowdsourced noise only); higher $\Delta p$ simulates increasingly degraded annotation quality. Experiments are conducted on D4RL locomotion tasks; the same $\Delta p$-fraction of non-tie labels is corrupted identically across all compared methods.

\begin{figure}[t]
  \centering
  \includegraphics[width=\columnwidth]{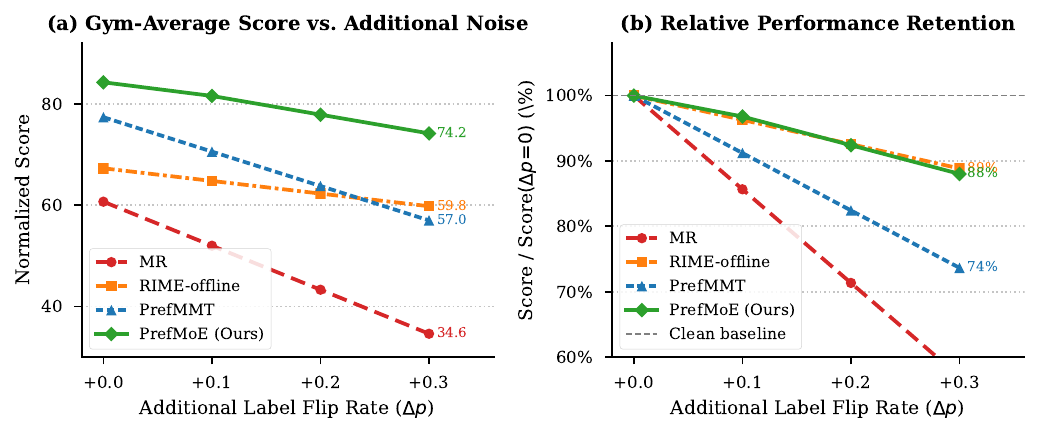}
  \caption{
    \textbf{Noise robustness on D4RL locomotion (Gym average).}
    $\Delta p$ denotes the \emph{additional} label flip rate applied on top of the 100-annotator natural label noise.
    \textbf{(a)}~Absolute Gym-average score.
    \textbf{(b)}~Relative retention: score normalised by each method's $\Delta p\!=\!+0$ performance.
  }
  \label{fig:noise_robustness}
  \vspace{-15pt}
\end{figure}

Fig.~\ref{fig:noise_robustness} visualizes both the absolute scores and the relative performance retention across $\Delta p$ levels.

\textbf{Low additional noise ($\Delta p \leq +0.1$):} PrefMoE and PrefMMT dominate. Under the crowdsourced baseline ($\Delta p\!=\!+0$), PrefMoE leads RIME-offline by $17$ and PrefMMT by $10$, suggesting that temporal sequence modeling
provides the primary advantage when additional synthetic corruption is absent.

\textbf{High additional noise ($\Delta p \geq +0.2$):} RIME-offline overtakes PrefMMT. Each additional $0.1$ flip rate reduces PrefMMT by $\sim\!6.8$ on average, while RIME-offline loses only $\sim\!2.5$ per increment. RIME's KL trust filter dynamically excludes samples whose ensemble prediction disagrees strongly with the stored label; this mechanism, designed for corrupted labels, becomes increasingly effective as $\Delta p$ grows. PrefMMT absorbs further erroneous gradient signals from the added flips. At $\Delta p\!=\!+0.2$, RIME-offline first exceeds PrefMMT; at $\Delta p\!=\!+0.3$, RIME-offline clearly exceeds PrefMMT ($59.8$ vs.\ $57.0$).

\textbf{PrefMoE across all $\Delta p$ levels:}
PrefMoE degrades at only $\sim\!3.4$ per $0.1$ increment, slower than PrefMMT at $6.8$ and approaching RIME-offline at $2.5$, maintaining a consistent $14$ to $17$ absolute lead over RIME-offline throughout. This implicit noise resistance arises from the soft routing mechanism: when preference pairs carry corrupted labels, the router can assign them higher weight to a general-purpose expert, limiting contamination of experts that have specialised on clean preference modes. At $\Delta p\!=\!+0.3$, PrefMoE retains $88\%$ of its baseline performance while maintaining a lead, whereas PrefMMT retains only $74\%$. These results suggest that PrefMoE combines the best properties of both axes: temporal sequence modeling dominates at low additional noise, and expert routing provides structural noise resilience as label quality degrades, without requiring the explicit denoising infrastructure of RIME.

\subsubsection{Does the MoE Architecture Scale Better as Annotator Pool Diversity Grows}

To understand how the size of the crowdsourced annotation pool affects each method, we vary the number of annotators whose labels are included in training: $N_{\text{ann}} \in \{10, 25, 50, 100\}$. A smaller pool produces more homogeneous labels with lower inter-annotator disagreement; a larger pool introduces greater diversity and natural label noise. All other settings follow the main experiment; no additional synthetic noise is applied ($\Delta p\!=\!+0$).

\begin{figure}[t]
  \centering
  \includegraphics[width=\columnwidth]{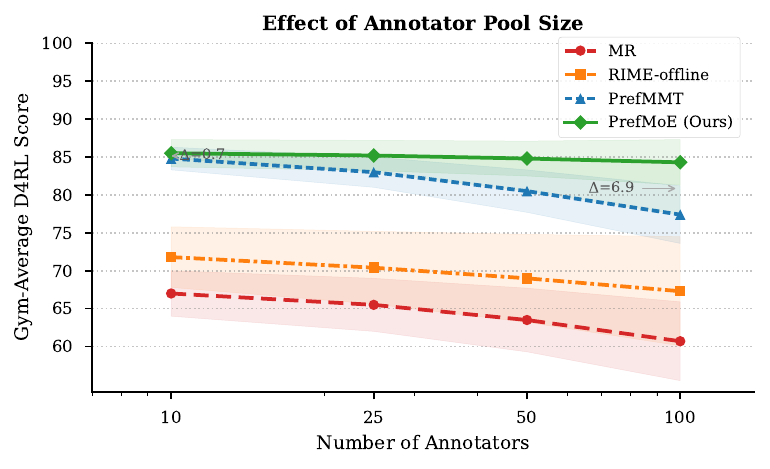}
  \caption{\textbf{Effect of annotator pool size on D4RL Gym-average score.} All conditions use the same total number of preference pairs. Shaded bands show $\pm$std over five seeds. The annotated $\Delta$ values mark the PrefMoE--PrefMMT gap at $N_{\text{ann}}\!=\!10$ and $N_{\text{ann}}\!=\!100$. PrefMoE is nearly insensitive to pool size, while PrefMMT degrades sharply as diversity grows. MR and RIME-offline degrade modestly.}
  \label{fig:annotator_ablation}
  \vspace{-15pt}
\end{figure}

Fig.~\ref{fig:annotator_ablation} reveals an asymmetry in how each method responds to growing annotation diversity. PrefMoE is nearly insensitive to pool size, maintaining a consistently high score as $N_{\text{ann}}$ increases from 10 to 100. This flatness reflects the design of the expert routing mechanism: additional annotators introduce new preference structure that the router absorbs by assigning it to underutilised experts, rather than forcing it to override existing specialisations.

\textbf{PrefMMT} degrades most sharply, as the single inter-modal encoder must simultaneously represent an increasingly wide range of conflicting evaluation criteria; with more annotators contributing divergent judgements, the encoder is pulled in competing directions and can no longer faithfully capture any individual preference mode. RIME-offline also degrades notably: its denoising filter targets per-label corruption rather than the systematic distributional diversity that arises when annotators genuinely hold different preferences, so it provides limited protection as the pool grows. MR shows a similar trend; its Markovian reward structure does not provide immunity to annotation diversity, and the growing label heterogeneity pushes its per-step reward estimates away from any consistent criterion.

Critically, the PrefMoE--PrefMMT gap is negligible at $N_{\text{ann}}\!=\!10$, where annotator preferences are relatively homogeneous and a single encoder suffices, but widens substantially by $N_{\text{ann}}\!=\!100$. This scaling behaviour is precisely the property required in realistic crowdsourced settings: the MoE architecture becomes relatively more advantageous as the annotation pool grows toward the scale typical of production RLHF deployments.

\section{Conclusion and Future Work}
We introduced PrefMoE, a MoE reward learning framework for PbRL under heterogeneous, crowdsourced supervision. By allowing multiple experts to specialize in different latent preference criteria and combining them through trajectory-level routing, PrefMoE models diverse and partially conflicting preference signals that a single reward model cannot represent. Experiments show that this design improves the robustness of preference prediction and leads to more reliable downstream policy learning. More broadly, our results highlight the importance of explicitly modeling preference diversity as large-scale datasets increasingly rely on crowdsourced or synthetic supervision. Future work includes integrating PrefMoE with explicit noise filtering, learning adaptive expert allocation, and extending the framework to online PbRL and large-scale alignment settings.

\typeout{}
\bibliography{main}
\bibliographystyle{IEEEtran}
\end{document}